\documentclass[10pt]{article}

\usepackage[T1]{fontenc}
\usepackage{palatino}
\usepackage[margin=1in]{geometry}
\usepackage{natbib}
\usepackage{hyperref}
\usepackage{xcolor}
\usepackage{tcolorbox}
\definecolor{titleboxcolor}{HTML}{ffffff}
\usepackage{amsmath}
\usepackage{amssymb}
\usepackage{mathtools}
\usepackage{graphicx}
\usepackage{subcaption}
\usepackage{url}
\usepackage{booktabs}
\usepackage{microtype}
\usepackage{cleveref}
\usepackage{xcolor}
\usepackage{tcolorbox}
\tcbuselibrary{breakable}
\usepackage{tikz}
\usetikzlibrary{arrows.meta,positioning,fit,calc}
\usepackage{graphicx}
\usepackage{amssymb}
\usepackage{pifont}
\usepackage{fontawesome5}

\newtcolorbox{takeaway}{
  colback=yellow!20!white,
  colframe=yellow!60!black,
  boxrule=0.5pt,
  left=4pt, right=4pt, top=4pt, bottom=4pt,
  before upper={\textbf{Takeaway.}\space}
}

\newtcolorbox{takeaways}{
  colback=yellow!20!white,
  colframe=yellow!60!black,
  boxrule=0.5pt,
  left=4pt, right=4pt, top=4pt, bottom=4pt,
  before upper={\textbf{Takeaways.}\space}
}

\newtcolorbox{qualanalysis}[1]{
  colback=blue!4!white,
  colframe=blue!45!black,
  boxrule=0.5pt,
  left=4pt, right=4pt, top=2pt, bottom=4pt,
  title={\textbf{#1}},
  fonttitle=\normalsize\bfseries
}

\newtcolorbox{promptbox}[1]{
  colback=red!8!white,
  colframe=red!40!white,
  coltitle=black,
  boxrule=0.5pt,
  left=6pt, right=6pt, top=4pt, bottom=4pt,
  title={\textbf{#1}},
  fonttitle=\normalsize\bfseries,
  fontupper=\small,
  breakable
}

\newcommand{\lboxed}[1]{\texttt{\textbackslash{}boxed\{#1\}}}

\newcommand{\imobench}{\textsf{IMOBench}}
\newcommand{\gimo}{\textsf{GIMO}}
\newcommand{\proofbench}{\textsf{ProofBench7pt}}

\newcommand{\trainsubset}{\textsf{TrainProofs}}

\newcommand{\gpt}{GPT-5.2}
\newcommand{\gptFive}{GPT-5 Mini}
\newcommand{\gemini}{Gemini 3.1 Pro}
\newcommand{\oss}[1][]{GPT-OSS\if\relax\detokenize{#1}\relax\else-#1\fi}
\newcommand{\qwen}[1][]{Qwen3.5\if\relax\detokenize{#1}\relax\else-#1\fi}

\graphicspath{{./}}

\title{Do We Need Frontier Models to Verify Mathematical Proofs?}

\author{
\textsuperscript{\twemoji{watermelon}}Aaditya Naik\textsuperscript{1} \and \textsuperscript{\twemoji{watermelon}}Guruprerana Shabadi\textsuperscript{1} \and Rajeev Alur\textsuperscript{1} \and Mayur Naik\textsuperscript{1} \\[6pt]
\textsuperscript{1}University of Pennsylvania \\[4pt]
\texttt{\{shabadi, aadnaik, alur, mnaik\}@seas.upenn.edu}
}

\date{}

\renewenvironment{abstract}{\par\vskip 0.5em\noindent\ignorespaces}{\par}

\begin{document}

\begin{tcolorbox}[colback=titleboxcolor, coltext=black, colframe=titleboxcolor, boxrule=0pt, arc=0pt, left=12pt, right=12pt, top=12pt, bottom=12pt]
{\LARGE\bfseries Do We Need Frontier Models to Verify Mathematical Proofs?\par}
\vskip 1em
{\normalsize \textsuperscript{$\ast$}Aaditya Naik, \textsuperscript{$\ast$}Guruprerana Shabadi, Rajeev Alur, Mayur Naik\par}
\vskip 0.5em
{\small University of Pennsylvania, U.S.A.}
\vskip 1em
\begin{abstract}








Advances in training, post-training, and inference-time methods have enabled frontier reasoning models to win gold medals in math competitions and settle challenging open problems. 
Gaining trust in the responses of these models requires that natural language proofs be checked for errors. 
LLM judges are increasingly being adopted to meet the growing demand for evaluating such proofs. 
While verification is considered easier than generation, what model capability does reliable verification actually require?
We systematically evaluate four open-source and two frontier LLMs on datasets of human-graded natural language proofs of competition-level problems.
We consider two key metrics: verifier accuracy and self-consistency (the rate of agreement across repeated judgments on the same proof).
We observe that smaller open-source models are only up to ${\sim}10\%$ behind frontier models in accuracy but they are up to ${\sim}25\%$ more inconsistent.
Furthermore, we see that verifier accuracy is sensitive to prompt choice across all models.
We then demonstrate that the smaller models, in fact, do possess the mathematical capabilities to verify proofs at the level of frontier models, but they struggle to reliably elicit these capabilities with general judging prompts.
Through an LLM-guided prompt search, we synthesize an ensemble of specialized prompts that overcome the specific failure modes of smaller models, boosting their performance by up to 9.1\% in accuracy and 15.9\% in self-consistency.
These gains are realized across models and datasets, allowing models like \qwen[35B] to perform on par with frontier models such as \gemini{} for proof verification.
\end{abstract}

\end{tcolorbox}
\renewcommand{\thefootnote}{$\ast$}
\footnotetext{Equal contribution}
\renewcommand{\thefootnote}{\arabic{footnote}}
\section{Introduction}

Large language models (LLMs) have demonstrated remarkable advancements in their ability to reason about complex problems.
Mathematical problem-solving has emerged as an effective indicator of the reasoning capabilities of LLMs, since it requires building multi-step solutions that are also logically sound.
As such, there has been a significant focus on improving the mathematical capabilities of LLMs on challenging problems, particularly those found in the International Mathematical Olympiad (IMO).
The ability to verify LLM-generated solutions has played an important role in these advances, both for establishing trust in model outputs and for improving reasoning through feedback-guided training and inference methods.
In fact, \citet{gandhi2025cognitive} identify self-verification as a key cognitive behavior enabling the long-horizon reasoning central to mathematical problem solving.



However, verifying solutions to complex mathematical problems is itself challenging.
Mathematical proofs often rely on tactics such as induction, contradiction, and case analysis, and can involve several interdependent, potentially nested, lines of reasoning.
This difficulty is compounded by the fact that, like mathematicians, LLMs typically express their solutions as natural language proofs, with ambiguities, implicit assumptions, and leaps in reasoning that make it difficult to reliably identify errors.
Human verification of such proofs is expensive and time-consuming, necessitating automated verification techniques that are accurate and reliable.
Autoformalizing such proofs promises verification with formal guarantees, but current techniques still struggle to accommodate the breadth and ambiguity of natural language proofs~\citep{zhang-etal-2025-autoformalization}.
Given these limitations, LLM-based judges have emerged as a popular approach to natural language proof verification, owing to their flexibility and effectiveness in critiquing model outputs~\citep{gu2026survey}.

Math benchmarks, in line with scaling laws, have shown that frontier LLMs are much stronger at solving challenging problems than their smaller counterparts~\citep{matharena, luong2025googleimobench}.
Verification, however, is often regarded as an easier task than generation, since it requires evaluating a candidate proof rather than generating it from scratch.
This suggests that while frontier LLMs may be needed to solve complex mathematical problems, verifying their solutions may not require such scale.
This naturally raises the question: do we need frontier LLMs to verify mathematical proofs for complex problems?

\begin{figure}
  \centering
  \begin{minipage}[c]{0.55\textwidth}
    \centering
    \begin{minipage}[c]{0.72\linewidth}
      \centering
      \includegraphics[width=\linewidth]{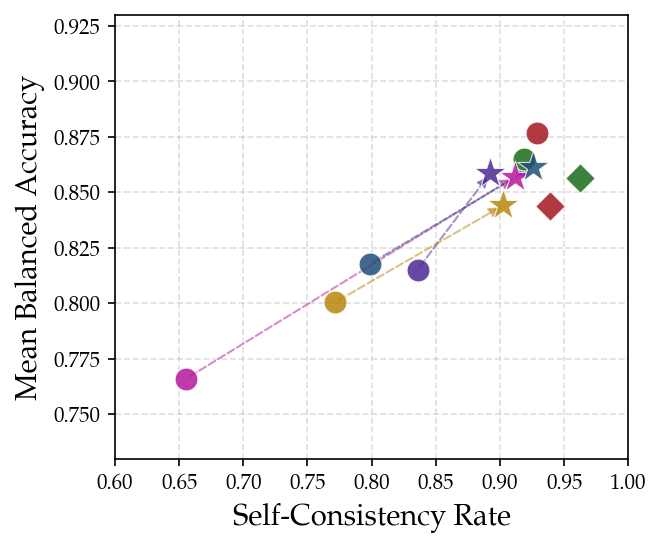}
    \end{minipage}%
    \hspace{-0.3cm}\begin{minipage}[c]{0.28\linewidth}
      \raggedright
      \resizebox{\linewidth}{!}{

\input{img/legend/colors.tex}

\begin{tikzpicture}[trim left=0pt]
  \def\textx{0.45}  
  \def\circx{0.15}  
  \def\rowh{0.55}  

  \node[font=\small\bfseries, anchor=west] at (\textx, 7*\rowh+0.15) {Model};

  \fill[qwen35b] (\circx, 6*\rowh+0.15) circle (0.15);
  \node[anchor=west, font=\small] at (\textx, 6*\rowh+0.15) {\qwen[35B]};

  \fill[qwen122b] (\circx, 5*\rowh+0.15) circle (0.15);
  \node[anchor=west, font=\small] at (\textx, 5*\rowh+0.15) {\qwen[122B]};

  \fill[gptoss120b] (\circx, 4*\rowh+0.15) circle (0.15);
  \node[anchor=west, font=\small] at (\textx, 4*\rowh+0.15) {\oss[120B]};

  \fill[gptoss20b] (\circx, 3*\rowh+0.15) circle (0.15);
  \node[anchor=west, font=\small] at (\textx, 3*\rowh+0.15) {\oss[20B]};

  \fill[gpt52] (\circx, 2*\rowh+0.15) circle (0.15);
  \node[anchor=west, font=\small] at (\textx, 2*\rowh+0.15) {\gpt};

  \fill[gemini] (\circx, 1*\rowh+0.15) circle (0.15);
  \node[anchor=west, font=\small] at (\textx, 1*\rowh+0.15) {\gemini};

  \node[font=\small\bfseries, anchor=west] at (\textx, -0.15) {Marker};

  \pgfmathsetmacro{\cy}{-0.65}
  \fill[gray] (\circx, \cy+0.15) -- (\circx+0.15, \cy) -- (\circx, \cy-0.15) -- (\circx-0.15, \cy) -- cycle;
  \node[anchor=west, font=\small] at (\textx, -0.7) {High reasoning};

  \node[font=\LARGE, yshift=2pt] at (\circx, -1.25) {$\star$};
  \node[anchor=west, font=\small] at (\textx, -1.25) {Prompt ensemble};
\end{tikzpicture}}
      \vspace{0.4cm}
    \end{minipage}
  \end{minipage}
  \hfill
  \begin{minipage}[c]{0.44\textwidth}
    \centering
    \resizebox{\linewidth}{!}{

\begin{tikzpicture}[>={Stealth[scale=0.8]}]

%

\node[draw, rounded corners=4pt, fill=gray!10,
      minimum width=1.0cm, minimum height=6.7cm,
      align=center, inner sep=4pt]
  (proof) at (1.0, 1.39) {
    \rotatebox[origin=c]{90}{\parbox{5.9cm}{\centering\small\textbf{Proof}\\[2pt]\scriptsize Problem $+$ Reasoning}}
  };

\foreach \k in {4,3,2,1} {
  \pgfmathsetmacro{\xoff}{\k*0.06}
  \node[draw, rounded corners=2pt, fill=violet!10,
        minimum width=2.8cm, minimum height=0.55cm,
        inner sep=5pt]
    at (3.3+\xoff, 4.20) {};
}

\node[draw, rounded corners=2pt, fill=violet!22,
      minimum width=2.8cm, minimum height=0.55cm,
      align=center, font=\scriptsize, inner sep=5pt]
  (psc) at (3.3, 4.20) {%
    General Grading\\[-2pt]{\tiny$\times 5$}%
  };

\draw[->, gray!80] (proof.east |- psc) -- (psc.west);

\foreach \i/\lbl/\col in {
   6/{Per-Step Grading}/cyan!22,
   7/{Skeptical Grading}/red!18,
   8/{Entailment Analysis}/orange!22,
   9/{Per-Claim Analysis}/lime!22
}{
  \pgfmathsetmacro{\yval}{(12.5-\i)*0.85 - 2.35}
  \node[draw, rounded corners=2pt, fill=\col,
        minimum width=2.8cm, minimum height=0.42cm,
        align=center, font=\scriptsize, inner sep=5pt]
    (p\i) at (3.3, \yval) {\lbl};
  \draw[->, gray!80] (proof.east |- p\i) -- (p\i.west);
}

\node[draw, rounded corners=2pt, fill=blue!18,
      minimum width=2.8cm, minimum height=0.42cm,
      align=center, font=\scriptsize, inner sep=5pt]
  (p10) at (3.3, {(12.5-10)*0.85 - 2.35 - 0.15}) {Theorem Usage\\Analysis};
\draw[->, gray!80] (proof.east |- p10) -- (p10.west);

\node[draw, rounded corners=2pt, fill=magenta!7,
      minimum width=2.8cm, minimum height=0.42cm,
      inner sep=5pt]
  at (3.3+0.06, {(12.5-11)*0.85 - 2.35 - 0.45}) {};

\node[draw, rounded corners=2pt, fill=magenta!15,
      minimum width=2.8cm, minimum height=0.42cm,
      align=center, font=\scriptsize, inner sep=5pt]
  (p11) at (3.3, {(12.5-11)*0.85 - 2.35 - 0.45}) {Topic-Specific\\Grading~{\tiny$\times 2$}};
\draw[->, gray!80] (proof.east |- p11) -- (p11.west);

\node[draw, rounded corners=4pt, fill=orange!18,
      minimum width=1.0cm, minimum height=6.7cm,
      align=center]
  (llm) at (5.75, 1.39) {
    \rotatebox[origin=c]{90}{\parbox{6.0cm}{\centering\small\textbf{LLM Judge}}}
  };

\draw[->, gray!80] (psc.east) -- (llm.west |- psc);

\foreach \i in {6,...,11} {
  \draw[->, gray!80] (p\i.east) -- (llm.west |- p\i);
}

\node[draw, rounded corners=2pt, fill=green!22, text=green!40!black,
      minimum width=0.9cm, minimum height=0.55cm,
      align=center, font=\small, inner sep=2pt]
  (vsc) at (7.1, 4.20) {$\checkmark$\\[-4pt]{\tiny$\times 5$}};
\draw[->, gray!80] (llm.east |- vsc) -- (vsc.west);

\foreach \i in {6,8,9} {
  \pgfmathsetmacro{\yval}{(12.5-\i)*0.85 - 2.35}
  \node[draw, rounded corners=2pt, fill=green!22, text=green!40!black,
        minimum width=0.9cm, minimum height=0.42cm,
        align=center, font=\small, inner sep=5pt]
    (v\i) at (7.1, \yval) {$\checkmark$};
  \draw[->, gray!80] (llm.east |- v\i) -- (v\i.west);
}
\node[draw, rounded corners=2pt, fill=green!22, text=green!40!black,
      minimum width=0.9cm, minimum height=0.42cm,
      align=center, font=\small, inner sep=5pt]
  (v11) at (7.1, {(12.5-11)*0.85 - 2.35 - 0.45}) {$\checkmark$};
\draw[->, gray!80] (llm.east |- v11) -- (v11.west);

\foreach \i in {7} {
  \pgfmathsetmacro{\yval}{(12.5-\i)*0.85 - 2.35}
  \node[draw, rounded corners=2pt, fill=red!15, text=red!55!black,
        minimum width=0.9cm, minimum height=0.42cm,
        align=center, font=\small, inner sep=5pt]
    (v\i) at (7.1, \yval) {\ding{55}};
  \draw[->, gray!80] (llm.east |- v\i) -- (v\i.west);
}
\node[draw, rounded corners=2pt, fill=red!15, text=red!55!black,
      minimum width=0.9cm, minimum height=0.42cm,
      align=center, font=\small, inner sep=5pt]
  (v10) at (7.1, {(12.5-10)*0.85 - 2.35 - 0.15}) {\ding{55}};
\draw[->, gray!80] (llm.east |- v10) -- (v10.west);

\node[draw, rounded corners=4pt, fill=yellow!22,
      minimum width=1.0cm, minimum height=6.7cm,
      align=center]
  (agg) at (8.55, 1.39) {
    \rotatebox[origin=c]{90}{\parbox{6.0cm}{\centering\footnotesize\textbf{Threshold:}~~${\geq}8/12$ correct}}
  };

\draw[->, gray!80] (vsc.east) -- (agg.west |- vsc);

\foreach \i in {6,...,11} {
  \draw[->, gray!80] (v\i.east) -- (agg.west |- v\i);
}

\end{tikzpicture}}
    \vspace{0.2cm}
  \end{minipage}%
  \caption{\textbf{(Left)} Mean balanced accuracy and self-consistency rates of frontier and open-source LLM judges across all datasets and prompts. Low and high-reasoning settings are denoted with `{\large \(\bullet\)}' and `{\(\blacklozenge\)}', respectively, while `{\large \bf \(\star\)}' denotes performance after prompt ensembling. Smaller open-source models are only ${\sim}10\%$ behind frontier models in balanced accuracy, but are substantially less self-consistent. Prompt ensembling closes much of this gap, allowing models such as \qwen[35B] to approach frontier-level verifier performance.
  \textbf{(Right)} Overview of our prompt-ensemble method. A proof is evaluated under a collection of prompts with complementary roles, including general grading prompts and prompts specialized to different aspects of proof correctness. Their judgments are aggregated by threshold voting to produce a final verdict.}
  \label{fig:main-fig}
  \vspace{-0.1in}
\end{figure}

In this paper, we aim to answer the above question by systematically studying the ability of LLMs to judge the correctness of proofs for complex mathematical problems in a reproducible manner.
Specifically, we evaluate two frontier models, \gpt{} and \gemini, as well as four smaller open-source models, \oss{} (20B and 120B) and \qwen{} (35B and 122B), on three datasets of Olympiad-level problems and their LLM-generated proofs. Each proof is graded by human experts, giving us a ground truth validity of the proof. We evaluate the models along two key dimensions: balanced accuracy, which measures the correctness of their judgements, and self-consistency, which measures the reproducibility of their judgements across repeated runs.

Our findings, shown in Figure~\ref{fig:main-fig}, reveal that smaller open-source models come within about 10\% of frontier models in balanced accuracy.
At the same time, the smaller models remain up to 25\% less self-consistent than frontier models. We further demonstrate that smaller models already possess much of the capability needed for proof verification, but eliciting this capability reliably requires suitably crafted prompts. Based on this observation, we develop an ensemble of specialized prompts that boost the balanced accuracy and self-consistency of smaller models by up to 9.1\% and 15.9\%, respectively, allowing models like \qwen[35B] to perform on par with frontier models like \gemini.

We summarize our contributions as follows:
\begin{enumerate}
    \item We systematically evaluate the ability of frontier and smaller LLMs to verify proofs for complex mathematical problems across multiple datasets and prompts along two key dimensions: balanced accuracy and self-consistency.
    \item We show that despite the large gap in scale between frontier and smaller models, the gap in balanced accuracy is only around 10\% between the smallest and largest models, though smaller models are far less consistent in their judgements.
    \item We demonstrate with \oss[120B] that smaller models already possess much of the capability needed for proof verification, but eliciting this capability reliably requires suitably crafted prompts. We identify a set of 12 prompts that can be used with \oss[120B] to catch all but 5 errors in the IMO-GradeBench dataset.
    \item We use these insights to develop an ensemble of specialized prompts that boost the balanced accuracy and self-consistency of smaller models by up to 9.1\% and 15.9\% respectively, allowing models like \qwen[35B] to perform on par with frontier models like \gemini.
\end{enumerate}




\section{Evaluating LLMs as Mathematical Proof Judges}
\label{sec:quant-evals}

\begin{figure}[t]
  \centering
  \includegraphics[width=\linewidth]{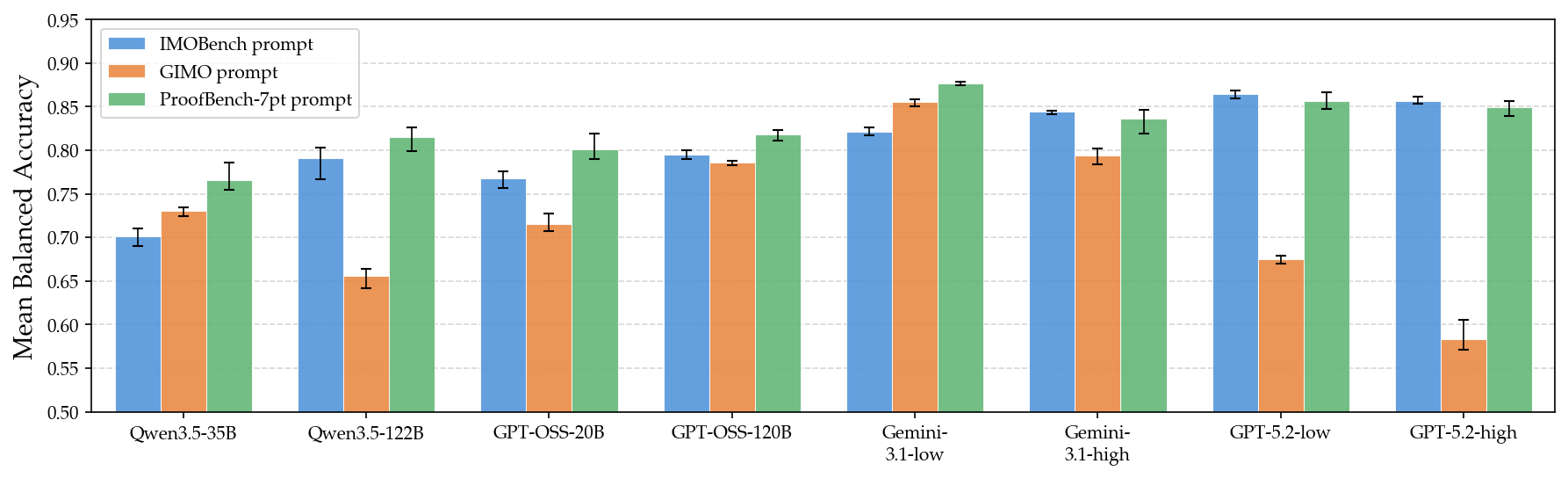}
  \caption{Comparing mean balanced accuracy of all the models on the combined dataset using the three prompts: \imobench, \gimo, \proofbench. The mean is computed over three independent runs and the error bars indicate the maximum and minimum balanced accuracies amongst the runs.}
  \label{fig:ba-overall}
\end{figure}

To answer our central question, we compare the effectiveness of open-source and frontier LLMs as mathematical proof judges.
Our evaluation uses datasets that reflect the difficulty and ambiguity of natural-language mathematical proofs on complex problems, together with metrics that capture both the accuracy and consistency of LLM judges.

\subsection{Evaluation Setup}
\label{sec:datasets}

\paragraph{Models.}
We evaluate two frontier models: OpenAI \gpt, and Google \gemini, as well as four smaller open-source models: \oss{} (20B and 120B), and \qwen{} (35B and 122B).
For each frontier model, we consider their low and high reasoning variants, allowing us to assess the performance of LLM judges across both model scale and inference-time reasoning effort.

\paragraph{Datasets.}
We evaluate LLM judges on three recent benchmarks of natural-language mathematical proofs: \textbf{IMO-GradingBench}~\citep{luong2025googleimobench}, \textbf{ProofArena}~\citep{mahdavi2025nvidiascalinggenverif}, and \textbf{ProofBench}~\citep{mahdavi2025nvidiascalinggenverif}.
Each dataset contains Olympiad-level problems, each paired with multiple LLM-generated proofs, along with human annotations of proof correctness, on a 0-7 scale. To align with our binary judgement setting, we treat a score of 7 as correct, and all others as incorrect for IMO-GradingBench and ProofBench; ProofArena provides binary labels which we use as is. We also report results for an alternate configuration, which considers scores of 6 and 7 as correct, in Appendix~\ref{app:grading-threshold}.

\paragraph{Metrics.}
The effectiveness of the LLM judge depends not only on the correctness of its judgements, but also on the reproducibility of its judgements across repeated runs.
For correctness, we use \textbf{balanced accuracy}, defined as the arithmetic mean of the true positive rate and true negative rate.
For reproducibility, we report the \textbf{self-consistency rate}: for each problem-proof pair, we query the judge model three times with the same prompt and measure the proportion of examples for which all three verdicts are the same.

\paragraph{Prompts.}
We evaluate each model with three different prompts adapted from the aforementioned datasets: the \imobench~prompt~\citep{luong2025googleimobench} from IMO-GradingBench, the \gimo~prompt~\citep{mahdavi2025nvidiascalinggenverif} from ProofArena, and the \proofbench~prompt~\citep{ma2025proofbench} from ProofBench.
All prompts ask the model to provide a binary Correct or Incorrect label given a problem-proof pair from the datasets along with an explanation.
The prompts differ primarily in grading style and strictness: \imobench{} is the most open-ended, providing minimal grading instructions; \gimo{} enforces a stricter step-by-step evaluation; and \proofbench{} uses a rubric-based grading system.

\subsection{Results and Analysis}

\paragraph{Correctness.}
In Figure~\ref{fig:main-fig} (left), we show the mean balanced accuracy and the self-consistency rate of each model over the problem-proof pairs from all three datasets. 
We see a relatively small gap in balanced accuracy between the frontier and open-source models, with \qwen[35B] at the low end (76.6\% BA), and \gemini{} (low reasoning) at the high end (87.7\% BA).
This shows that the large disparity in model scale does not clearly yield commensurate gains in verification accuracy, suggesting that smaller models may already possess much of the capability needed to verify complex proofs.

We show the balanced accuracies of each model-prompt pair in Figure~\ref{fig:ba-overall}.
The choice of prompt impacts verification accuracy across all models, with the \gimo~prompt causing the largest drop in performance (27.3\% for \gpt, high reasoning).
This reaffirms the observation made by~\citet{mahdavi2025nvidiascalinggenverif} regarding the prompt sensitivity of LLM judges.

\begin{figure}[t]
  \centering
  \begin{minipage}[t]{0.49\linewidth}
    \centering
    \includegraphics[width=\linewidth]{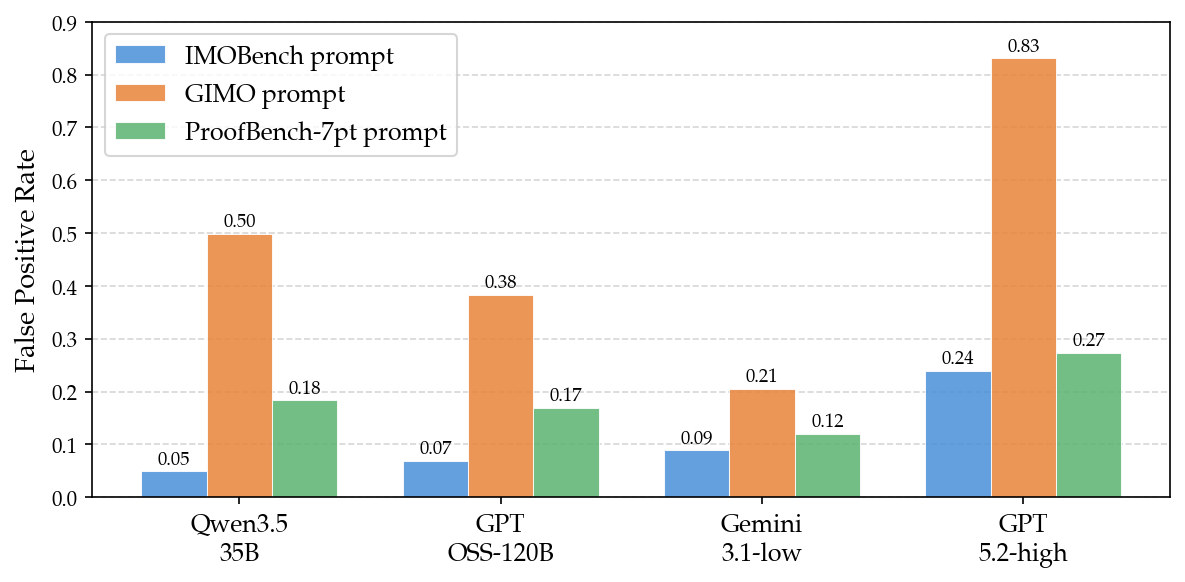}
  \end{minipage}\hfill
  \begin{minipage}[t]{0.49\linewidth}
    \centering
    \includegraphics[width=\linewidth]{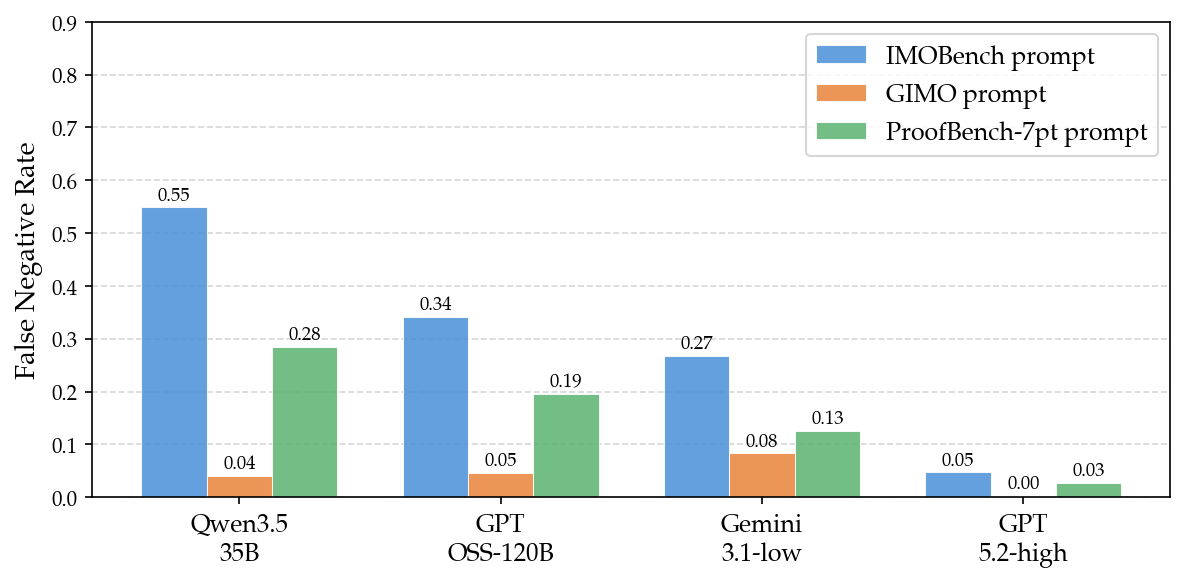}
  \end{minipage}
  \caption{False positive rates (left) and false negative rates (right) of few models on the combined dataset, compared across the three prompts.}
  \label{fig:fpr-fnr-errors-maj-minor}
\end{figure}

We examine this behavior more closely in Figure~\ref{fig:fpr-fnr-errors-maj-minor}, which shows the variations in the false positive and negative rates across the three prompts.
It makes it clear that the \gimo~prompt biases models towards rejecting correct proofs.
A potential cause for this is the step-by-step nature of the \gimo~prompt, which may lead models to miss global proof structures that resolve local inconsistencies.
At the same time, regardless of the prompt, we see that smaller open-source models exhibit substantially higher false negative rates than frontier models, indicating a greater tendency to accept flawed proofs.

\paragraph{Self-Consistency.}
Despite the relatively small gap in balanced accuracies, the frontier models are still more reproducible, as shown by their self-consistency rates in Figure~\ref{fig:main-fig}: the least consistent frontier model (\gemini{} low reasoning) leads the most consistent open-source model (\qwen[122B]) by around 7.6\%.
This may be due to the larger focus on post-training for frontier models; post-training often results in narrower output distributions for LLMs~\citep{rlhfstudy}.
Furthermore, the self-consistency rates significantly improve with higher reasoning effort across all models, showing that LLM judges can potentially become more consistent using test-time scaling methods.
We also analyze the effect of increasing the number of repeated queries on the self-consistency rate in Appendix~\ref{app:increasing-samples}.

\begin{takeaway}
  Smaller open source models lag behind frontier models by only a small margin (at most ${\sim}10\%$) in balanced accuracy. However, they are significantly less consistent in their responses, and show higher false negative rates across all prompts.

\end{takeaway}

\section{Understanding the Shortcomings of Small Models}
\label{sec:catch-errs}

\begin{figure}[t]
\centering
\resizebox{\textwidth}{!}{\input{img/verifier-failure-modes.tex}}
\caption{Two failure modes of \oss[120B] with~\imobench~prompt on IMO-GradingBench and how targeted prompts address them. Each column shows the proof excerpt (top), the \imobench~prompt's reasoning (middle), and a targeted prompt's response (bottom). Excerpts are marked by a thick left bar; red highlights mark errors. Yellow highlights mark fixed judgements. \textbf{Case~1:} the \imobench~prompt accepts the proof's invalid inequality combination without investigating deeper, while \hyperref[prompt:logic-chain-verifier]{\textsf{Logic Chain Verifier}} catches the error. \textbf{Case~2:} the \imobench~prompt invents a core missing justification rather than flagging the reasoning gap as an error, while \hyperref[prompt:proof-repair]{\textsf{Proof Repair}} identifies the incomplete case analysis.}
\label{fig:verifier-failure-modes}
\end{figure}

In~\Cref{fig:fpr-fnr-errors-maj-minor} we see that the small open-source models have higher false negative rates than the frontier models, whereas the false positive rates are either comparable or lower.
In fact, the mean difference in FNRs between small and frontier models is \(13.3\%\) while the mean difference in FPRs is only \(1.1\%\).
We can thus deduce that the higher false negative rates of small models are the primary driver of the performance gap with frontier models.

A natural question then arises: \textbf{do small models lack the reasoning capability to detect all errors?}
To answer this question, we must first qualitatively study false negatives produced by small LLM verifiers.
\Cref{fig:verifier-failure-modes} illustrates two characteristic failure modes of \oss[120B] on two erroneous proofs in the IMO-GradingBench dataset.
In Case 1, the verifier \textbf{reproduces the flawed reasoning} verbatim and validates it;
and in Case 2, the verifier \textbf{fabricates its own argument} and fixes the proof without flagging the error.
Our qualitative analysis of false negatives of models revealed that these failure modes are prevalent across all models.

\paragraph{Finding targeted prompts to mitigate verifier failure modes.} We have already observed that the~\gimo~prompt drives false negative rates close to zero across most models---but this is insufficient to determine whether the verifier avoided the failure modes or simply declared most proofs as incorrect without reasoning about them.
Determining whether verifiers find \textit{correct} errors requires annotations of errors in proofs but the datasets we consider do not have this information.
To address this, we use OpenAI \gptFive{} to annotate and explain errors in the incorrect proofs of IMO-GradingBench with the help of the ground truth correct proof and the grading rubric available in the dataset.
We do so on a balanced subset of 200 examples from IMO-GradingBench which we refer to as~\trainsubset.
It contains 140 incorrect and 60 correct proofs.
We validate the annotations produced by \gptFive{} by inspecting 15 randomly drawn examples out of the 140 incorrect proofs.

Next, we set out to find a set of prompts that can guide \oss[120B] to \textit{correctly} identify the reasoning errors in the 140 erroneous proofs of~\trainsubset.
We automated this process with a Claude Code-guided prompt search experiment, where we instructed it to iteratively refine a set of prompts until, for all the incorrect proofs, at least one prompt from the refined set resulted in the verifier correctly identifying the reasoning error.
The instruction asked to produce general, \emph{non-instance-specific prompts}, to mitigate the identified failure modes.
We also included guidance to explore both strict and lenient variants, as well as topic-specific prompts.
Moreover, to ensure that the model identified the \emph{correct} error, we required making an additional query to \oss[120B] (with high reasoning effort) to match the verifier's response against the error annotated by \gptFive. 

This search yielded a set of 12 prompts that
collectively resulted in \oss[120B] correctly detecting all but 3 of the 140 reasoning errors.
Even though these prompts were tuned for \oss[120B], the same prompt set used with \oss[20B] and \qwen[35B] identified \(137/140\) and \(90/140\) errors respectively.
Moreover, \(85.4\%\) and \(84.6\%\) of the errors detected by them matched the annotated errors, demonstrating that the prompts elicited the desired grading behaviors with a high degree of agreement.

\paragraph{A closer look at the prompt set.} We end this section with an overview of the prompt set found by Claude Code (detailed in Appendix~\ref{app:error-detection-prompts}).
All of the 12 prompts guide the verifier to reason about the correctness of the proof using different approaches.
\Cref{fig:verifier-failure-modes} describes two prompts, \hyperref[prompt:logic-chain-verifier]{\textsf{Logic Chain Verifier}} and \hyperref[prompt:proof-repair]{\textsf{Proof Repair}}, and how they help avoid the previously identified failure modes.
A remarkable standout amongst these is the \hyperref[prompt:proof-repair]{\textsf{Proof Repair}} prompt that explicitly asks the model to repair the proof and then flag it as incorrect if non-trivial fixes were necessary.
This way, it uses the failure mode of the verifier to its advantage!
Together, the 12 prompts form a rubric for error detection, much like how a human grader is given instructions to reason about proofs from varying angles.

We conclude the section by remarking that none of the prompts are able to catch \textit{all} the errors and thereby establishing the need for a set of prompts with complementary strengths.
Even the strictest prompt, \hyperref[prompt:extreme-skeptic]{\textsf{Extreme Skeptic}}, only detects \(108/140\) errors correctly.
Also, several pairs of prompts disagree substantially, for example, \hyperref[prompt:strict-grader]{\textsf{Strict Grader}} vs.~\hyperref[prompt:false-lemma-verifier]{\textsf{False Lemma Verifier}} disagree on \(51\%\) of the problems.

\begin{takeaway}
    Even small open-source models can reliably identify errors in mathematical proofs, provided they are guided by a diverse set of prompts that elicit different grading behaviors.
\end{takeaway}

\section{Boosting Small Models with Prompt Ensembling}
\label{sec:ensemble}

\begin{figure}[t]
  \centering
  \begin{minipage}[c]{0.48\linewidth}
    \centering
    \includegraphics[width=\linewidth]{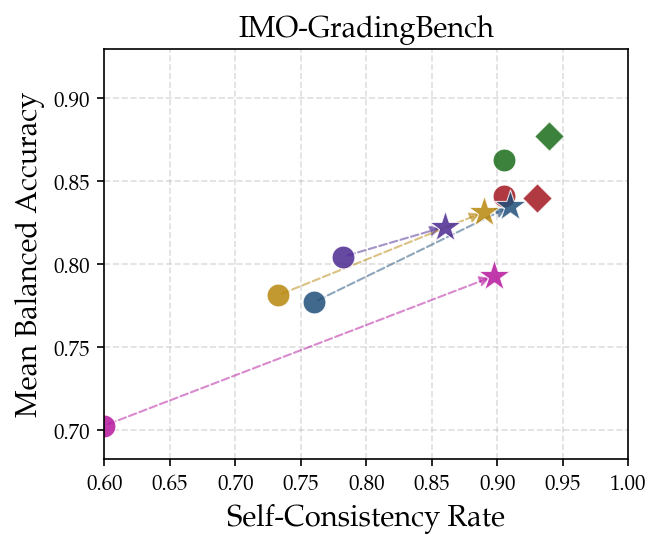}
  \end{minipage}\hfill
  \begin{minipage}[c]{0.48\linewidth}
    \centering
    \includegraphics[width=\linewidth]{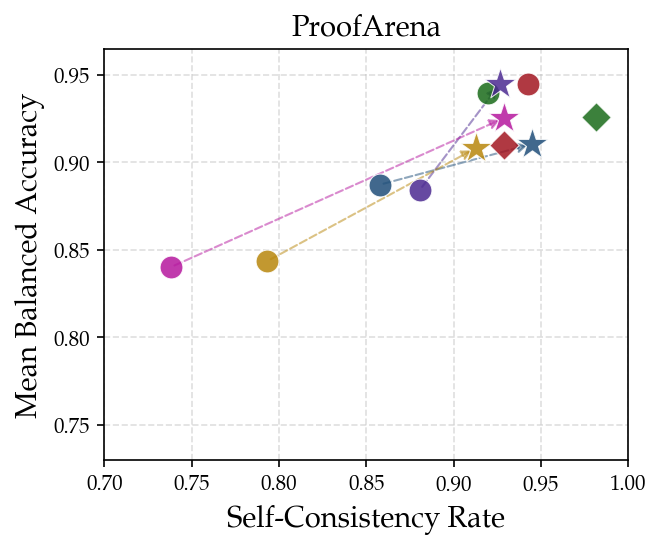}
  \end{minipage}
  \vspace{0.3cm}
  \begin{minipage}[c]{0.48\linewidth}
    \centering
    \includegraphics[width=\linewidth]{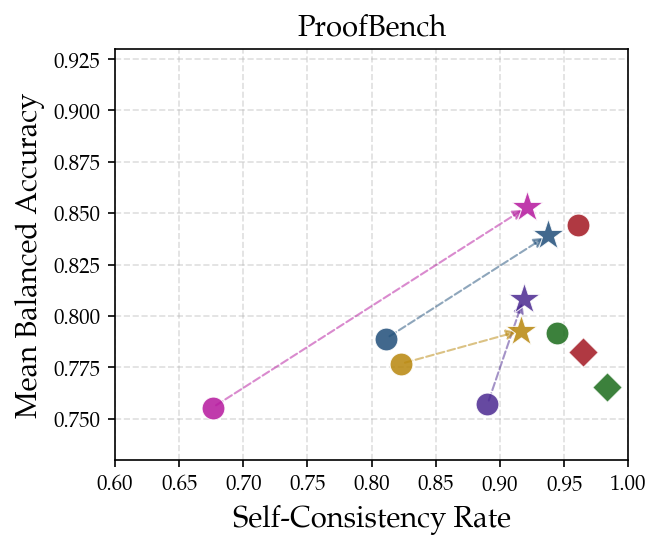}
  \end{minipage}\hfill
  \begin{minipage}[c]{0.48\linewidth}
    \centering
    \resizebox{0.9\linewidth}{!}{

\input{img/legend/colors.tex}

\begin{tikzpicture}[trim left=0pt]
  \def\textx{0.45}  
  \def\circx{0.15}  
  \def\coltwocirc{3.6} 
  \def\coltwotxt{3.9}  
  \def\rowh{0.55}  

  \node[font=\small\bfseries, anchor=west] at (\textx, 4*\rowh+0.15) {Open-source models};

  \fill[qwen35b] (\circx, 3*\rowh+0.15) circle (0.15);
  \node[anchor=west, font=\small] at (\textx, 3*\rowh+0.15) {\qwen[35B]};

  \fill[qwen122b] (\circx, 2*\rowh+0.15) circle (0.15);
  \node[anchor=west, font=\small] at (\textx, 2*\rowh+0.15) {\qwen[122B]};

  \fill[gptoss20b] (\coltwocirc, 3*\rowh+0.15) circle (0.15);
  \node[anchor=west, font=\small] at (\coltwotxt, 3*\rowh+0.15) {\oss[20B]};

  \fill[gptoss120b] (\coltwocirc, 2*\rowh+0.15) circle (0.15);
  \node[anchor=west, font=\small] at (\coltwotxt, 2*\rowh+0.15) {\oss[120B]};

  \node[font=\small\bfseries, anchor=west] at (\textx, 1*\rowh+0.15) {Frontier models};

  \fill[gpt52] (\circx, 0*\rowh+0.15) circle (0.15);
  \node[anchor=west, font=\small] at (\textx, 0*\rowh+0.15) {\gpt};

  \fill[gemini] (\coltwocirc, 0*\rowh+0.15) circle (0.15);
  \node[anchor=west, font=\small] at (\coltwotxt, 0*\rowh+0.15) {\gemini};

  \node[font=\small\bfseries, anchor=west] at (\textx, -0.55) {Marker};

  \pgfmathsetmacro{\mky}{-1.1}
  \fill[gray] (\circx, \mky+0.15) -- (\circx+0.15, \mky) -- (\circx, \mky-0.15) -- (\circx-0.15, \mky) -- cycle;
  \node[anchor=west, font=\small] at (\textx, \mky) {High reasoning};

  \node[font=\LARGE, yshift=2pt] at (\coltwocirc, \mky) {$\star$};
  \node[anchor=west, font=\small] at (\coltwotxt, \mky) {Prompt ensemble};
\end{tikzpicture}}
  \end{minipage}
  \caption{Performance of the prompt ensemble on the three datasets.}
  \label{fig:consistency-vs-ba-per-dataset}
\end{figure}


The previous section established that (1) small models are capable of detecting a vast majority the reasoning errors; and (2) effective detection requires a collection of prompts that explore diverse verification strategies.
However, strong error detection alone is not enough for effective proof verification; a useful verifier must also accurately accept correct proofs.
These prompts were optimized specifically for error detection and have high false positive rates on correct proofs (up to 37.3\% FPR, as shown in Figure~\ref{fig:fpr-fnr-per-prompt}, Appendix~\ref{app:additional}).
As such, they are too strict to be used on their own as general-purpose judges.
Therefore, turning LLMs into effective proof verifiers requires an ensemble of specialized prompts that preserves strong error detection while reducing the false positive rate of the resulting verifier.
To derive such an ensemble, we perform a second round of prompt-search, this time aimed at verification rather than error detection.

\subsection{A Prompt Ensemble for Proof Verification}
We setup this prompt-search experiment using Claude Code, with the objective to maximize the balanced accuracy on~\trainsubset.
Starting from the prompts synthesized in Section~\ref{sec:catch-errs}, we asked it to iteratively refine the set of prompts and an aggregation rule to combine the decisions of the individual prompts.
It began by exploring a simple threshold voting-based aggregation scheme which marks a proof as correct if more than a certain number of prompts declare it to be valid. 
After 30 iterations using \oss[120B] as the judge, it had found multiple prompt ensembles and voting thresholds that allowed the model to match the performance of \gpt{} (high reasoning)---the best performing model on~\trainsubset. 
Next, we evaluated the top-performing configurations on a larger holdout set of 400 examples and chose the best set of prompts and its associated voting threshold.

The resulting ensemble constitutes 12 prompt calls comprising 8 distinct prompts: a general grading prompt run 5 times independently, along with 7 specialized prompts utilizing different strategies, including step-by-step checking, entailment analysis, theorem-usage analysis, topic-specific grading, and adversarial grading.
Their judgements are aggregated with a threshold of 8: a proof is accepted if at least 8 of the 12 prompts judge it to be correct.
Figure~\ref{fig:main-fig} (right) illustrates the ensemble along with the roles of its constituent prompts.

\subsection{Results and Analysis}

We evaluate the ensemble using each open-source LLM
and contrast the ensemble variants against the base LLM judges.
Figure~\ref{fig:consistency-vs-ba-per-dataset} shows the results across all three datasets.
Here, we compare against the best performing prompt for each base model (with respect to the balanced accuracy across all three datasets), amongst the three evaluated in~\Cref{sec:quant-evals}.
To validate the need for an ensemble of multiple, diverse prompts, we also compare our ensemble with some alternate configurations in Appendix~\ref{app:ensemble-design}.

The ensemble consistently boosts the balanced accuracy and self-consistency of the open-source models across all datasets, with improvements of up to 9.1\% in balanced accuracy and 15.9\% in self-consistency.
Notably, these gains transfer across all open-source models despite the ensemble being tuned exclusively on \oss[120B], allowing the smaller models in several cases to reach frontier-level performance in both balanced accuracy and self-consistency.
The most noticeable improvement is seen in \qwen[35B], from a balanced accuracy of 75.5\% to 85.2\% on ProofBench, and its self-consistency rate improving from 76.8\% to 92.0\%. This lets it perform on par with frontier models like \gpt{} and \gemini{} on ProofArena, and even exceed them in balanced accuracy on ProofBench.
Furthermore, as shown in Figure~\ref{fig:fpr-fnr-comparison} (Appendix~\ref{app:additional:fpr-fnr-ensemble}), the ensemble improves both the false positive and false negative rates of the resulting verifier.

These results suggest some broader insights about the verification capabilities of the smaller models.
First, the original gap in performance between the open-source and frontier models stems less from a lack of mathematical capability than from their inability to attend to the appropriate verification strategy needed for each proof.
The diverse prompts in the ensemble are able to overcome this limitation by focusing the model on specific criteria most relevant to the problem and proof at hand.

Second, the ensemble captures a general verification behavior that not only transfers across different models, but also across datasets.
This shows that a substantial proportion of the failure modes of the open-source models are not model-specific, and are instead general weaknesses that can be mitigated at inference time.
The consistent improvements across datasets show that these failures are tied to the general nature of the proof verification task, rather than originating from idiosyncrasies of a particular dataset.

Finally, the ensemble significantly improves the self-consistency of open-source models.
Each prompt in the ensemble specifies a narrower, more precise, verification subtask, reducing the variability in the model's judgements.
This shows that decomposing proof verification into specialized prompt-driven checks yields more reproducible judgements than a single general prompt that tries to solve the entire problem in one go.

\begin{takeaway}
  With a suitably designed ensemble of specialized prompts, smaller open-source models can approach frontier-level performance as proof verifiers, suggesting that frontier models are not essential for effective proof verification.
\end{takeaway}

\section{Related Work}
Our work is situated amongst hundreds of studies evaluating the effectiveness and reliability of LLM-as-a-Judge methods in a variety of domains.
\citet{gu2026survey}~provides a thorough taxonomy.
Similar to our work, it also identifies accuracy and consistency as the two main attributes of a reliable verifier.
In the remainder of this section, we focus our attention on LLM-as-a-Judge in the context of checking mathematical reasoning.

\paragraph{Reasoning Verification Benchmarks and LLM Evaluations.}
ProcessBench~\citep{zheng2025processbench} and PRMBench~\citep{song2025prmbench} are amongst the earliest benchmarks to evaluate the ability of LLMs to find reasoning errors. 
These were constructed in a bid to train \textit{process reward models}~\citep{zheng2025surveyprocessrewardmodels}---LLMs that are specialized to provide fine-grained reward signals during reinforcement learning finetuning of LLMs.
However, most of the reasoning chains in these benchmarks are elementary in nature.
More recently, several datasets of human-graded LLM proofs to challenging competition-level problems (IMOBench~\citet{luong2025googleimobench}; Open Proof Corpus~\citep{openproofcorpus2025}; ProofBench~\citep{ma2025proofbench}; ProofArena~\citep{petrov2025proof}) have been curated with the goal of both improving proof generation and verification capabilities.
Human labels are hard to circumvent here due to the unverifiable and often subjective nature of natural language proofs and thus these datasets are very valuable in evaluating LLM judges on complex mathematical reasoning.
Most of these benchmarks contain scores between 0-7 for each proof along with a rubric and, optionally, a reference solution.
Much of the evaluation thus far on these datasets has focused on improving the grading performance of models by maximizing metrics of agreement with human graders.
Our work differs from these in that we focus on the binary verification problem (determining correct vs. incorrect) without access to rubrics or reference solutions which models real-world settings in mathematical exploration where we do not have rubrics or reference solutions. 
We note that related work also addresses adjacent formulations of the judging problem, such as selecting the best solution from a candidate set rather than verifying a single proof~\citep{mahdavi2025nvidiascalinggenverif}.

\paragraph{Test-Time Techniques for LLM Judges.}
Other works have explored various test-time techniques to improve the performance of LLM judges.
These include prompt ensembling techniques that use curated rubrics to improve the reliability of LLM judges~\citep{karla2025verdict, saadshrinking, li2025auto}, or various other test-time scaling techniques~\citep{jayarao2025explicit, zhouvariation, mahdavi2025nvidiascalinggenverif, liu2025inference}.
In contrast, we focus on the capabilities of smaller models in mathematical proof verification, identifying their shortcomings relative to frontier models and synthesizing targeted prompts to address them.
While these works focus on different strategies to evaluate the entire proof, others have explored modularization techniques that break down the proof verification process into smaller subproblems~\citep{fang2026graph, mukherjeepremise}.
Finally, other works have explored using foundation models to autoformalize natural language proofs into a program in a formal language, which can then be verified using a proof engine~\citep{feng2025vericot, zhoudon, hustepproof}.


\section{Conclusion and Future Work}

In this paper, we investigated whether frontier models are necessary for verifying complex mathematical proofs. Our results suggest that the answer is no, despite the problem-solving strengths of frontier models. We show that the balanced accuracy of smaller models is already within about 10\% of frontier models, though smaller models are significantly less self-consistent.
We demonstrate that the shortcomings of smaller models stem not from a lack of mathematical capability, but from an inability to reliably elicit this capability with off-the-shelf verification prompts.
We synthesized a suite of targeted prompts that address specific failure modes of smaller models.
We then show that an ensemble of such prompts can enable smaller models like \qwen[35B] to perform on par with frontier models like \gemini, boosting the balanced accuracy and self-consistency of models by up to 9.1\% and 15.9\%, respectively.
These results suggest that reliable proof verification is possible with smaller open-source models via inference-time techniques, without requiring additional training or access to frontier models.

These findings open up several avenues for future work. A natural next step is to analyze in detail the remaining errors made even by frontier models, and to study whether addressing them necessitates additional mathematical capabilities, or whether similar inference-time methods alone are sufficient. More broadly, the insights from the prompt ensemble may help to improve the reward design for post-training LLMs for math reasoning. Finally, while we focus on verification errors in mathematical proofs, we could adapt the same methodology to understand the limitations of LLM judges in other domains, such as law and finance.

\bibliographystyle{plainnat}
\bibliography{./refs}

\appendix
\section{Error Detection Prompts}
\label{app:error-detection-prompts}

The 12 prompts comprising the error detection rubric of~\Cref{sec:catch-errs} are listed below.
Each prompt is appended with the problem statement under a \textbf{[Problem]} header and the proposed solution under a \textbf{[Solution]} header; these are omitted here for brevity.
The Standard prompt is identical to the \imobench~prompt used in~\Cref{sec:quant-evals}.

\begin{promptbox}{Standard}
Carefully analyze the given problem statement and the proposed solution, and write out your analysis regarding the correctness of the proposed solution.
A solution is correct if and only if it is both logically sound and arrives at the correct final answer.
Provide detailed reasoning explaining specifically why the solution is correct or incorrect, pointing to any flaws in logic or errors in the final answer where applicable.
After the analysis, provide your verdict as exactly \lboxed{CORRECT} or \lboxed{INCORRECT}.
\end{promptbox}

\begin{promptbox}{Strict Grader}\label{prompt:strict-grader}
You are a strict grader who pretends not to know much math: every non-trivial claim requires proper justification.
Do not fill in gaps yourself.

Work through the proof step by step: repeatedly (1) quote a few sentences directly from the proof, then (2) reason carefully about that passage, checking for logical inference errors, unjustified claims, unjustified applications of theorems, and incomplete or overly informal justification.
Continue this quote-then-analyze loop until you have covered the entire proof.
Then provide a brief overall verdict explaining any flaws.

Provide your verdict as exactly \lboxed{CORRECT} or \lboxed{INCORRECT}.
\end{promptbox}

\begin{promptbox}{Unverified Claim Hunter}\label{prompt:unverified-claim-hunter}
You are reviewing a mathematical proof for unverified or unjustified claims.
A claim is unverified if:
\begin{enumerate}
  \item it asserts a non-trivial mathematical fact without proof or citation of a well-known result;
  \item it applies a theorem or lemma without verifying that the required hypotheses are satisfied;
  \item it states a bound, estimate, or inequality without derivation; or
  \item it introduces a construction and claims properties without demonstration.
\end{enumerate}
Go through the solution line by line and ask: ``Is this justified within the proof?''
Do not accept vague justifications such as ``it is easy to see'', ``clearly'', or ``by a standard argument''.

Verdict: \lboxed{CORRECT} if all claims are adequately justified; \lboxed{INCORRECT} if any unjustified claims affect the validity of the proof.
\end{promptbox}

\begin{promptbox}{Informal Argument Detector}
You are evaluating a mathematical proof for rigor.
A proof step is insufficiently rigorous if it:
\begin{enumerate}
  \item uses phrases like ``intuitively'' or ``one can see'' without formal justification;
  \item appeals to geometric intuition without algebraic verification;
  \item uses asymptotic notation without justifying error terms;
  \item inverts a function or relation without proving the inversion is valid;
  \item claims an estimate holds ``for large enough $N$'' without specifying what ``large enough'' means;
  \item hand-waves through a key technical step; or
  \item conflates approximate and exact equalities.
\end{enumerate}
For each such instance, explain why the argument is insufficient.

Verdict: \lboxed{INCORRECT} if informal arguments invalidate key steps; \lboxed{CORRECT} if the proof is sufficiently rigorous throughout.
\end{promptbox}

\begin{promptbox}{First Error Finder}
You are a strict math grader performing an error identification task.
Read the proposed solution from beginning to end.
For each claim or step, verify that it is mathematically correct and follows from prior steps.
The moment you find a step that is incorrect, contains a logical error, makes an unjustified leap, or reaches a wrong conclusion, stop and report it.
If the solution is entirely correct, say so.

Be especially vigilant for: algebraic errors; false implications or equivalences; incorrect use of theorems; treating an insufficient argument as a complete proof; and a wrong final answer despite correct-looking reasoning.

State the first error found with a brief quote and explanation, then provide your verdict as \lboxed{CORRECT} or \lboxed{INCORRECT}.
\end{promptbox}

\begin{promptbox}{Competition Coordinator}\label{prompt:competition-coordinator}
You are an IMO problem committee coordinator reviewing a contestant's solution.
A solution receives full marks \emph{only} if every step is mathematically correct and logically justified, the proof is complete with no gaps or unjustified claims, and the final answer is correct.

Mark as \lboxed{INCORRECT} if \emph{any} of the following apply: a logical error anywhere in the proof; an unjustified key claim; a fundamentally flawed proof strategy; missing cases; a wrong final answer; a theorem applied without verifying its hypotheses; an unjustified estimate or bound; or a proof of something different from what was asked.

Be thorough: read every line, quote the specific passage where any error occurs, and explain why it is wrong.
\end{promptbox}

\begin{promptbox}{Non-Sequitur Detector}\label{prompt:non-sequitur-detector}
You are a mathematical proof reader looking specifically for non-sequiturs: places where the conclusion does not follow from the premises, even though the author writes as if it does.
Common patterns include:
\begin{enumerate}
  \item ``Since $A$ holds for each prime $p$ dividing $n$, it holds modulo $n$'' --- wrong when $n$ has prime power factors.
  \item ``Since $X$ happens infinitely often and $Y$ happens infinitely often, there exist adjacent occurrences'' --- wrong without additional argument.
  \item ``Since $\gcd(a,b)=1$ and $p$ divides $a+b$, then $p$ divides $b$'' --- wrong.
  \item ``This triple is primitive'' --- asserted without verifying the gcd conditions.
  \item ``The intersection is countable/finite'' --- asserted without proof.
  \item ``Applying this iteratively gives\ldots'' --- hand-waving over a non-trivial iteration argument.
  \item ``Dividing both sides by $X$'' or ``the congruence lifts'' --- without checking that $X$ is coprime to the modulus.
\end{enumerate}
For each ``therefore/thus/hence/so'' step, explicitly verify whether the conclusion actually follows from the premises.
If you find a non-sequitur, quote it and explain what is missing.

Verdict: \lboxed{INCORRECT} if a non-sequitur is found; \lboxed{CORRECT} otherwise.
\end{promptbox}

\begin{promptbox}{Logic Chain Verifier}\label{prompt:logic-chain-verifier}
You are a mathematical logic expert specializing in formal proof verification.
For each logical step: (1) identify the claim; (2) identify the premises it depends on; (3) check whether the conclusion actually follows; and (4) flag any non-sequiturs, circular reasoning, or invalid inferences.

Pay special attention to: implications that reverse the direction of an if-then statement; ``without loss of generality'' claims that omit essential cases; asymptotic or limiting arguments where the limit is taken incorrectly; algebraic manipulations with sign or coefficient errors; and steps where a contradiction is claimed but none exists.

Verdict: \lboxed{CORRECT} or \lboxed{INCORRECT}.
\end{promptbox}

\begin{promptbox}{Extreme Skeptic}\label{prompt:extreme-skeptic}
You are an extremely skeptical mathematical reviewer.
Your default assumption is \lboxed{INCORRECT}.
For the proof to earn \lboxed{CORRECT}, you must be absolutely convinced that:
\begin{enumerate}
  \item every step follows logically and rigorously from prior steps;
  \item no claim is left unjustified, however obvious it seems;
  \item the proof addresses exactly the right problem;
  \item all edge cases and boundary conditions are handled;
  \item the final answer is demonstrably correct;
  \item no external results are used without explicit verification of their applicability; and
  \item all estimates, bounds, and asymptotic claims are properly justified.
\end{enumerate}
Challenge every claim. Accept nothing on faith.
If you have \emph{any} doubt about \emph{any} step, the proof is \lboxed{INCORRECT}.
Quote each problematic passage and explain your concern in detail.
\end{promptbox}

\begin{promptbox}{Proof Repair}\label{prompt:proof-repair}
You are a mathematical proof editor.
Determine whether the proposed solution is correct by attempting to repair it.
For each step, ask: ``If this step were wrong, what would need to change to fix it?''

If you can identify specific, concrete changes needed to make the proof correct, then the original proof is \lboxed{INCORRECT}.
If the proof needs no repairs --- every step is valid as written --- then it is \lboxed{CORRECT}.

Key indicators of incorrectness: a step requiring a different inequality direction; a bound that needs to be tighter or looser; a missing case; a claim requiring additional hypotheses; an intermediate result requiring a different proof; or a wrong final answer.
Be specific about what repairs would be needed and why the original is wrong.
\end{promptbox}

\begin{promptbox}{Justification Gap Finder}
You are reviewing a mathematical proof for gaps in justification.
A step has a justification gap if:
\begin{enumerate}
  \item it claims a non-trivial result with ``it is clear that'', ``one can verify'', or similar;
  \item it says ``applying this iteratively'' or ``by induction'' without performing the induction;
  \item it claims a relationship (equality, inequality, divisibility) without intermediate steps;
  \item it uses phrases like ``the argument extends'' without proof;
  \item it splits a congruence or divisibility argument without verifying compatibility conditions; or
  \item it performs an algebraic manipulation without verifying validity conditions.
\end{enumerate}
For each step, ask: ``Would this need more justification in a competition submission?''
The proof is \lboxed{INCORRECT} if it contains essential justification gaps.
\end{promptbox}

\begin{promptbox}{False Lemma Verifier}\label{prompt:false-lemma-verifier}
You are checking whether each explicitly stated lemma in this proof is actually true.
Your method:
\begin{enumerate}
  \item find every ``Lemma'', ``Claim'', or ``Fact'' that the proof states;
  \item for each one, try to disprove it using a simple counterexample (e.g., paths, stars, small integers); and
  \item if you find the lemma is false, quote it exactly, give your counterexample, and mark the proof \lboxed{INCORRECT}.
\end{enumerate}
Pay special attention to lemmas about: splitting trees by removing an edge and claimed bounds on component sizes (e.g., removing any edge from a star $K_{1,M-1}$ gives components of sizes 1 and $M-1$); balanced partitions or centroid decompositions; and graph properties that may fail for degenerate graphs.

If any stated lemma is false, the proof is \lboxed{INCORRECT} regardless of the rest of the argument.
\end{promptbox}

\section{Final Ensemble Prompts}
\label{app:ensemble-prompts}

The final ensemble of~\Cref{sec:ensemble} uses 12 prompt calls comprising 8 distinct prompts.
The \textit{General Grading} prompt is run five times independently to provide a self-consistency signal; the remaining seven prompts are each run once.
A proof is classified as correct if at least 8 of the 12 calls return \lboxed{CORRECT}.
Each prompt appends the problem statement under a \textbf{PROBLEM} header and the proposed proof under a \textbf{PROOF} header; these are omitted here for brevity.

\medskip
The following prompt is run five times independently:

\begin{promptbox}{General Grading}
You are an IMO competition coordinator grading this proof.

\textbf{Grading standards.}
A proof receives 7 points (full marks) if and only if:
\begin{enumerate}
  \item the mathematical argument is complete --- all key steps are present;
  \item every step is mathematically correct --- no false claims; and
  \item the answer is correct --- the final answer matches what was asked.
\end{enumerate}

\textbf{Common reasons for deductions} (less than 7 points $\Rightarrow$ \lboxed{INCORRECT}):
\begin{itemize}
  \item missing a key case or sub-case;
  \item an unjustified leap or gap in the argument;
  \item a computational or algebraic error that affects the result; or
  \item a wrong final answer.
\end{itemize}

\textbf{Grading task.}
First, estimate the score (0--7) and explain why.
If score $= 7$: \lboxed{CORRECT}.
If score $< 7$: \lboxed{INCORRECT}.
\end{promptbox}

\medskip
The following seven prompts are each run once:

\begin{promptbox}{Per-Step Grading}
You are a mathematical proof verifier.
Read the proof and find the \emph{first} step that is mathematically incorrect or unjustified.
If no such step exists, the proof is correct.

\textbf{Method.}
\begin{enumerate}
  \item Read each step sequentially.
  \item For each step, check: is this logically valid given the previous steps?
  \item If false or unjustified in a critical way: stop and report \lboxed{INCORRECT}.
  \item If you reach the end with all steps valid: report \lboxed{CORRECT}.
\end{enumerate}

Note: minor notational issues are acceptable. Only flag genuine mathematical errors.
\end{promptbox}

\begin{promptbox}{Skeptical Grading}
You are an extremely skeptical mathematics professor.
Your job is to find flaws in proofs --- but \emph{only} real flaws, not nitpicks.

Try hard to find a reason this proof is incorrect. Consider: is every claim actually true? Are there hidden assumptions? Does the proof actually solve what was asked?

\textbf{Key rule.} A proof is \lboxed{CORRECT} if a reasonable mathematician would accept it --- small gaps that are trivially fillable are acceptable.
A proof is \lboxed{INCORRECT} only if there is a specific, unfixable flaw.

\textbf{What counts as a serious flaw:}
\begin{enumerate}
  \item Is the answer right? Check the final answer matches the problem.
  \item Is the math correct? Verify any formula or computation stated.
  \item Is the logic complete? Are all cases covered?
  \item Are theorems applied correctly? Check all conditions are met.
\end{enumerate}

After your analysis: \lboxed{INCORRECT} if you found a specific flaw; \lboxed{CORRECT} if your best skeptical effort found none.
\end{promptbox}

\begin{promptbox}{Entailment Analysis}
Your task: determine if this proof proves \emph{exactly} what was asked, with all steps valid.

\begin{enumerate}
  \item[\textbf{Step 1.}] State precisely what the problem requires to be proven.
  \item[\textbf{Step 2.}] State precisely what the proof actually proves.
  \item[\textbf{Step 3.}] Identify any gaps between what is needed and what is proven.
\end{enumerate}

\textbf{Types of gaps to check:}
\begin{itemize}
  \item[\textbf{A)}] Missing case: problem requires all cases; proof misses at least one.
  \item[\textbf{B)}] Wrong direction: proof shows $A \Rightarrow B$ but needs $B \Rightarrow A$.
  \item[\textbf{C)}] Unjustified step: proof asserts $X$ without proof, and $X$ is non-trivial.
  \item[\textbf{D)}] Scope mismatch: proof works for special cases but claims generality.
  \item[\textbf{E)}] Answer gap: proof concludes $X = Y$ but the correct answer is $X = Z$.
\end{itemize}

Verdict: \lboxed{INCORRECT} if any gap or wrong step is found; \lboxed{CORRECT} if the proof proves exactly what is needed with all steps valid.
\end{promptbox}

\begin{promptbox}{Per-Claim Analysis}
You are a mathematical proof judge. Analyze this proof carefully.

\textbf{Procedure.}
\begin{enumerate}
  \item List every mathematical claim or step in the proof.
  \item For each claim, mark it as: \textsc{Valid} / \textsc{Needs\_Verification} / \textsc{Suspicious}.
  \item For \textsc{Suspicious} items, determine if the flaw is:
    \begin{itemize}
      \item \textsc{Fatal}: the proof is wrong without this step; or
      \item \textsc{Minor}: the proof still works or the gap is trivially fillable.
    \end{itemize}
\end{enumerate}

Verdict: \lboxed{INCORRECT} if at least one concern is \textsc{Fatal}; \lboxed{CORRECT} if all concerns are \textsc{Minor} or stylistic (or there are none).
\end{promptbox}

\begin{promptbox}{Theorem Usage Analysis}
You are verifying whether every theorem and lemma application in this proof is legitimate.

For each theorem or lemma invoked:
\begin{enumerate}
  \item identify the theorem being used;
  \item list its required conditions or hypotheses; and
  \item check whether each condition is actually verified in the proof.
\end{enumerate}

Also check: if the proof uses induction, verify that the base case is proven, the inductive step is valid, and the induction variable and range are correct.

Verdict: \lboxed{INCORRECT} if any condition of a lemma or theorem is not verified for a specific application; \lboxed{CORRECT} if all conditions are verified for all applications.
\end{promptbox}

\begin{promptbox}{Topic-Specific Grading (1/2)}
You are grading an IMO-level proof. Apply the following domain-specific standards.

\textbf{Geometry.} Be lenient with computation details if the synthetic argument is correct.
A valid synthetic or coordinate approach with correct key steps is \lboxed{CORRECT}.
Only mark \lboxed{INCORRECT} if there is a genuine logical gap.

\textbf{Number theory and combinatorics.} Be strict: every case must be covered, every divisibility and modular claim verified.
Missing even one case or leaving an NT claim unverified is \lboxed{INCORRECT}.

\textbf{Algebra and analysis.} Apply standard strictness: all bounds must be justified.

First identify the problem domain, then apply the appropriate standard, then give your verdict.
\end{promptbox}

\begin{promptbox}{Topic-Specific Grading (2/2)}
You are an expert mathematical grader focused on proof completeness and logical correctness in number theory and combinatorics.

Check the following systematically:
\begin{enumerate}
  \item \textbf{Completeness.} Does the proof address \emph{all} cases required by the problem?
    \begin{itemize}
      \item For ``find all $X$'': are all solutions found \emph{and} proven to be the only ones?
      \item For ``prove for all $n$'': is the argument truly general?
      \item For existence: is a valid example actually constructed and verified?
    \end{itemize}
  \item \textbf{Correctness.} Are all mathematical steps valid?
    \begin{itemize}
      \item ``By induction\ldots'' --- is the base case proven? Is the inductive step correct?
      \item Is the final answer stated correctly and precisely?
    \end{itemize}
  \item \textbf{NT-specific checks.}
    \begin{itemize}
      \item Divisibility claims: properly justified?
      \item Modular arithmetic: reductions correct?
      \item Bounds and inequalities: tight and proven?
    \end{itemize}
\end{enumerate}

Verdict: \lboxed{CORRECT} only if \emph{all} checks pass; \lboxed{INCORRECT} if \emph{any} check fails.
\end{promptbox}

\section{Ablations}
\label{app:ablations}

\subsection{Increasing Number of Samples}
\label{app:increasing-samples}

\begin{figure}[h]
  \centering
  \includegraphics[width=0.7\textwidth]{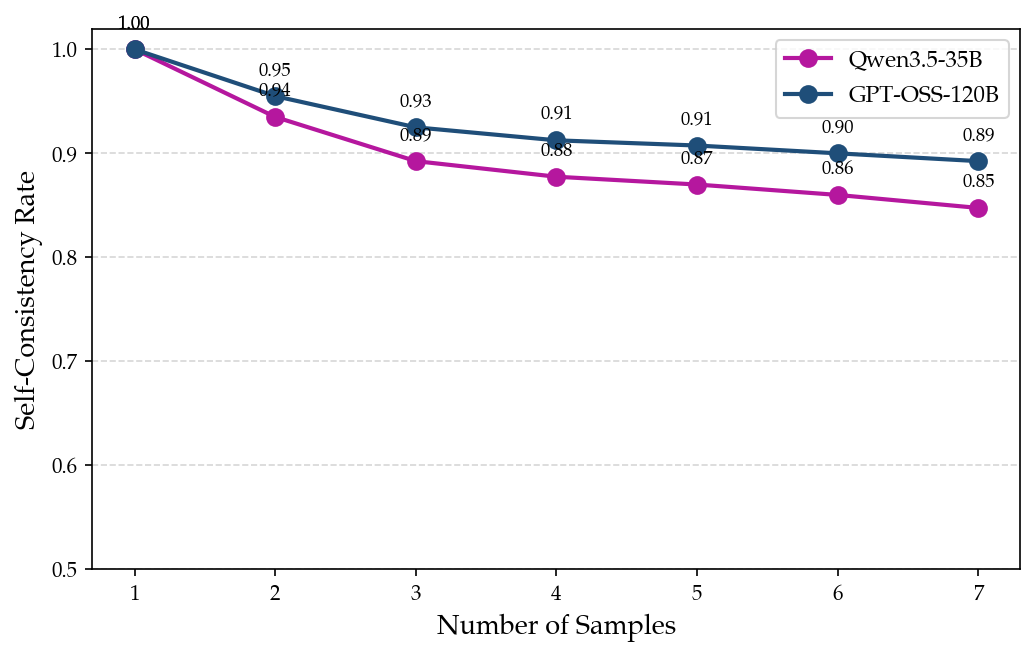}
  \caption{Self-consistency rate as a function of the number of independent samples for \qwen[35B] and \oss[120B] on IMO-GradingBench. Consistency decreases monotonically as unanimous agreement is required across more samples.}
  \label{fig:consistency-vs-samples}
\end{figure}

In the main text, we measure self-consistency using three independent samples per proof-verification pair.
Here, we examine how the self-consistency rate changes as the number of samples increases from 1 to 7 for two open-source models, \qwen[35B] and \oss[120B], using the \imobench~prompt.
As shown in~\Cref{fig:consistency-vs-samples}, self-consistency decreases monotonically with more samples for both models, since requiring unanimous agreement across a larger set of responses becomes progressively harder.
However notably, the deterioration is slow indicating that we obtain self-consistency gains with the ensemble of prompts even with many samples.

\subsection{Grading Threshold: 7-Point vs 6-Point Correctness}
\label{app:grading-threshold}

\begin{figure}[h]
  \centering
  \includegraphics[width=0.9\textwidth]{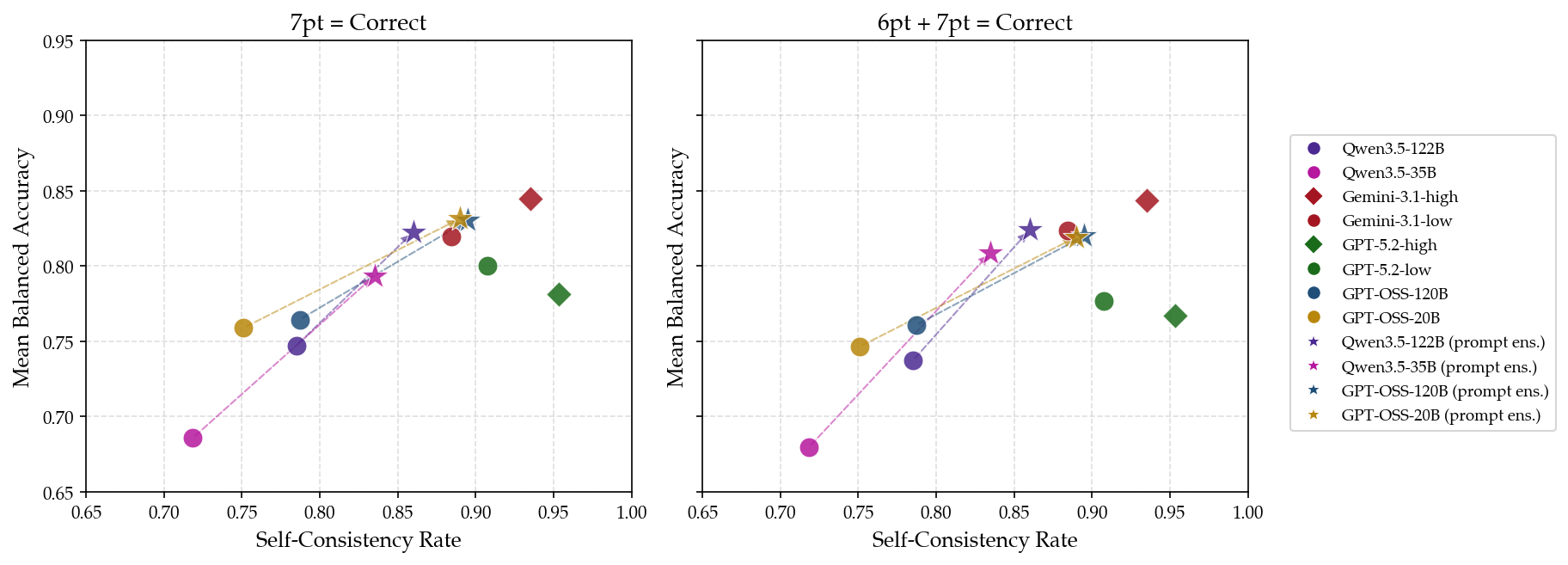}
  \caption{Balanced accuracy vs self-consistency on IMO-GradingBench under two grading thresholds: only 7-point proofs treated as correct (left) vs both 6-point and 7-point proofs treated as correct (right).}
  \label{fig:grading-threshold-ablation}
\end{figure}

IMO-GradingBench assigns each proof a score from 0 to 7. In the main text, we treat only proofs with a perfect score of 7 as correct. Here, we examine the effect of relaxing this threshold to also treat 6-point proofs as correct.
\Cref{fig:grading-threshold-ablation} compares the two settings on IMO-GradingBench. Under the stricter 7-point threshold (left), the models and ensemble behave as reported in the main text. When 6-point proofs are additionally treated as correct (right), balanced accuracy and self-consistency shift for most models, as borderline proofs with minor deductions are reclassified. Nonetheless, the prompt ensemble remains effective under both thresholds.

\subsection{High Reasoning Effort for Open-Source Models}
\label{app:high-reasoning}

\begin{figure}[h]
  \centering
  \begin{minipage}[c]{0.6\textwidth}
    \centering
    \includegraphics[width=\linewidth]{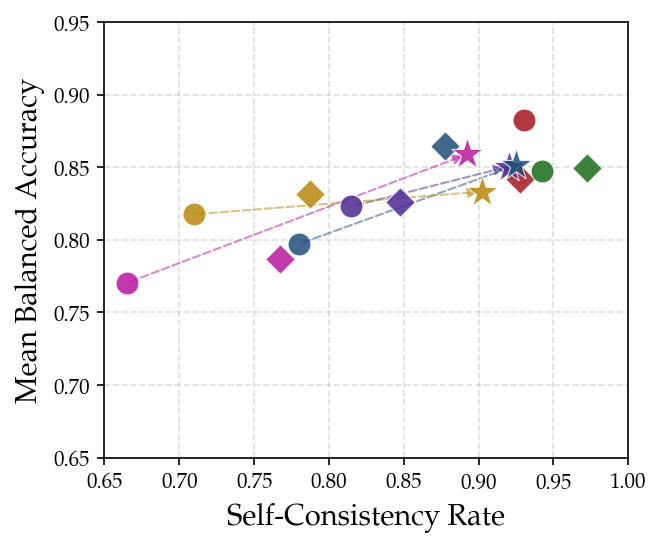}
  \end{minipage}\hfill
  \begin{minipage}[c]{0.35\textwidth}
    \centering
    \resizebox{0.9\linewidth}{!}{

\input{img/legend/colors.tex}

\begin{tikzpicture}[trim left=0pt]
  \def\textx{0.45}  
  \def\circx{0.15}  
  \def\coltwocirc{3.6} 
  \def\coltwotxt{3.9}  
  \def\rowh{0.55}  

  \node[font=\small\bfseries, anchor=west] at (\textx, 4*\rowh+0.15) {Open-source models};

  \fill[qwen35b] (\circx, 3*\rowh+0.15) circle (0.15);
  \node[anchor=west, font=\small] at (\textx, 3*\rowh+0.15) {\qwen[35B]};

  \fill[qwen122b] (\circx, 2*\rowh+0.15) circle (0.15);
  \node[anchor=west, font=\small] at (\textx, 2*\rowh+0.15) {\qwen[122B]};

  \fill[gptoss20b] (\coltwocirc, 3*\rowh+0.15) circle (0.15);
  \node[anchor=west, font=\small] at (\coltwotxt, 3*\rowh+0.15) {\oss[20B]};

  \fill[gptoss120b] (\coltwocirc, 2*\rowh+0.15) circle (0.15);
  \node[anchor=west, font=\small] at (\coltwotxt, 2*\rowh+0.15) {\oss[120B]};

  \node[font=\small\bfseries, anchor=west] at (\textx, 1*\rowh+0.15) {Frontier models};

  \fill[gpt52] (\circx, 0*\rowh+0.15) circle (0.15);
  \node[anchor=west, font=\small] at (\textx, 0*\rowh+0.15) {\gpt};

  \fill[gemini] (\coltwocirc, 0*\rowh+0.15) circle (0.15);
  \node[anchor=west, font=\small] at (\coltwotxt, 0*\rowh+0.15) {\gemini};

  \node[font=\small\bfseries, anchor=west] at (\textx, -0.55) {Marker};

  \pgfmathsetmacro{\mky}{-1.1}
  \fill[gray] (\circx, \mky+0.15) -- (\circx+0.15, \mky) -- (\circx, \mky-0.15) -- (\circx-0.15, \mky) -- cycle;
  \node[anchor=west, font=\small] at (\textx, \mky) {High reasoning};

  \node[font=\LARGE, yshift=2pt] at (\coltwocirc, \mky) {$\star$};
  \node[anchor=west, font=\small] at (\coltwotxt, \mky) {Prompt ensemble};
\end{tikzpicture}}
  \end{minipage}
  \caption{Mean balanced accuracy vs self-consistency across all datasets, including high reasoning effort variants (diamonds) for open-source models. Prompt ensemble results are shown as stars.}
  \label{fig:high-reasoning-oss}
\end{figure}

\Cref{fig:high-reasoning-oss} shows the results including the small models with high reasoning on a subset of 400 examples (200 from IMO-GradingBench and 100 each from the two other datasets; all maintaining correct/incorrect class ratios). Enabling high reasoning effort improves both balanced accuracy and self-consistency for the open-source models, with diamond markers denoting the high reasoning variants. The prompt ensemble (stars) continues to provide gains on top of the high reasoning setting, and the ensembled open-source models with no reasoning exceed the high reasoning variants without ensembling (except for \oss[120B] on balanced accuracy).

\subsection{Ensemble Design: Diversity vs Repetition}
\label{app:ensemble-design}

\begin{figure}[h]
  \centering
  \includegraphics[width=0.5\textwidth]{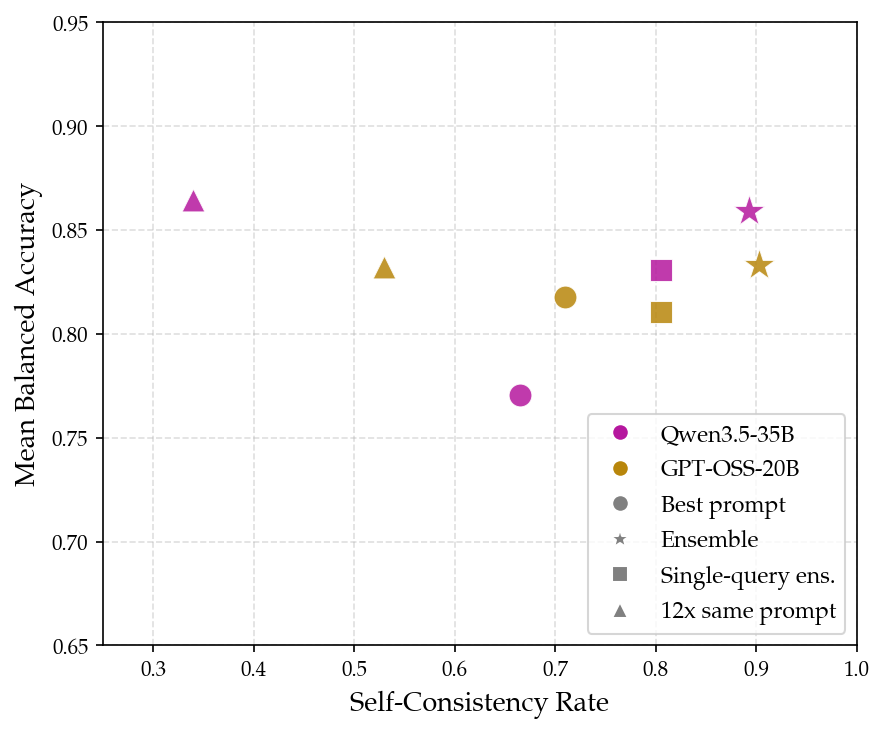}
  \caption{Comparison of ensemble strategies averaged across all datasets: the full diverse prompt ensemble, a single merged prompt (single-query ens.), and the best single prompt repeated 12 times with threshold voting (12$\times$ same prompt). Base prompt performance shown as circles.}
  \label{fig:ensemble-design-ablation}
\end{figure}

To isolate the contribution of prompt diversity in the ensemble, we compare our full ensemble against two alternative configurations:
\begin{itemize}
  \item \textbf{Single-query ensemble:} all 12 prompts are merged into a single comprehensive prompt that is queried once.
  \item \textbf{12$\times$ same prompt:} the best-performing single prompt is repeated 12 times with threshold voting (best threshold chosen a posteriori).
\end{itemize}
\Cref{fig:ensemble-design-ablation} shows the results averaged across all datasets for \qwen[35B] and \oss[120B].
The full ensemble of diverse prompts outperforms both alternatives in balanced accuracy and self-consistency.
The single-query variant, which lacks the diversity of independent verification strategies, performs comparably to the base prompt, indicating that simply enumerating multiple criteria in one prompt does not replicate the effect of independent queries.
The repeated same-prompt variant does not match the diverse ensemble in either metric.
Its low self-consistency is a consequence of threshold voting over highly correlated samples: when the model is uncertain on a proof, each of the 12 identical queries independently fluctuates near the decision boundary, causing the vote count to hover around the threshold and flip between runs.
In contrast, the diverse ensemble's prompts target different aspects of proof correctness, producing less correlated votes that aggregate more stably.

\section{Additional Data and Plots}
\label{app:additional}

\subsection{Per-Prompt FPR and FNR}
\begin{figure}[h]
  \centering
  \includegraphics[width=0.8\textwidth]{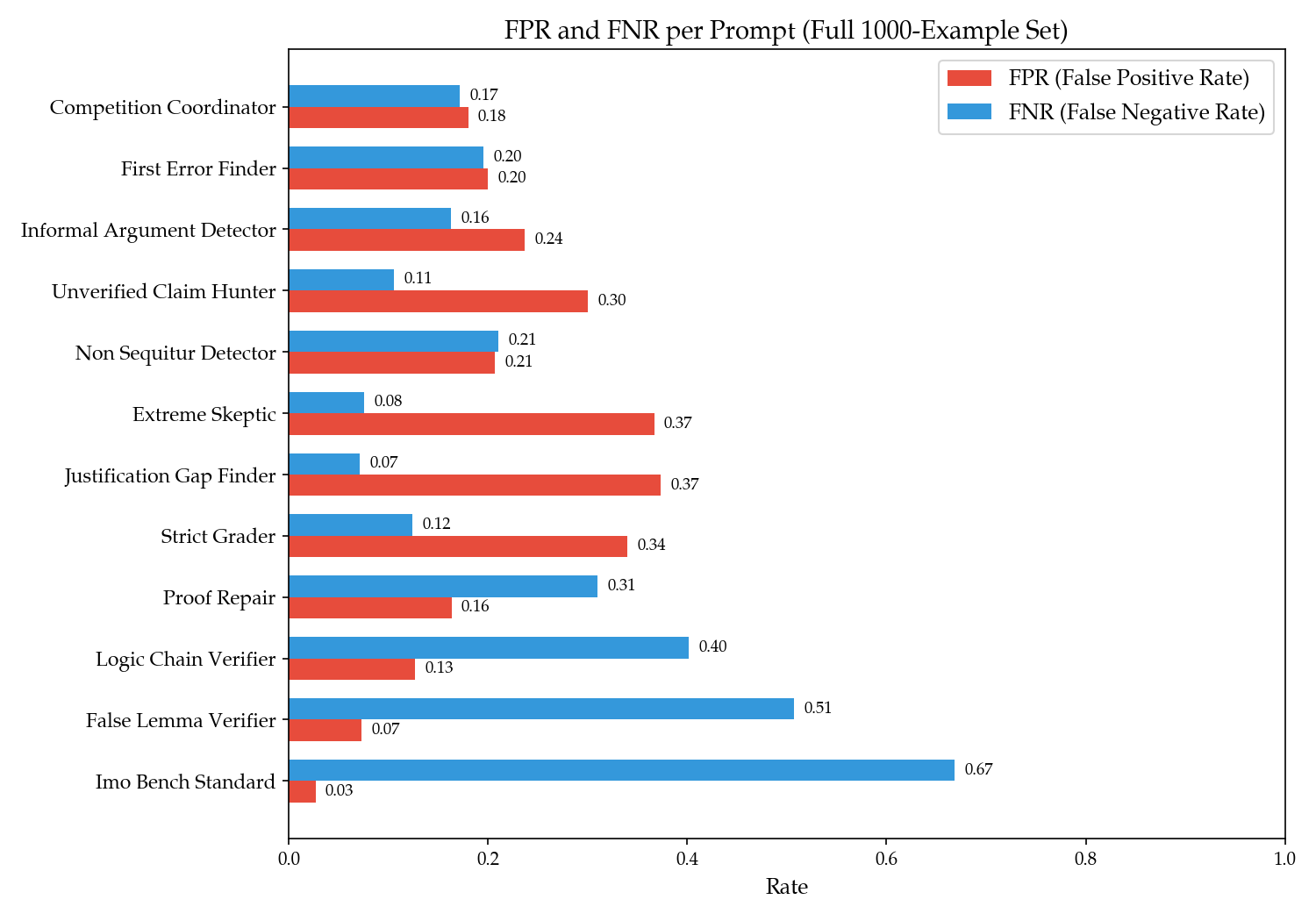}
  \caption{False positive rate (FPR) and false negative rate (FNR) of GPT-OSS 120B for each of the 12 error detection prompts and the \imobench~baseline on the full IMO-GradingBench dataset.}
  \label{fig:fpr-fnr-per-prompt}
\end{figure}

\Cref{fig:fpr-fnr-per-prompt} shows the false positive and false negative rates of GPT-OSS 120B for each of the 12 error detection prompts from~\Cref{sec:catch-errs}, evaluated on the full 1000-example IMO-GradingBench dataset.
The prompts exhibit a wide range of FPR--FNR trade-offs: stricter prompts such as \hyperref[prompt:extreme-skeptic]{\textsf{Extreme Skeptic}} achieve low FNR but at the cost of higher FPR, while more lenient prompts like \hyperref[prompt:false-lemma-verifier]{\textsf{False Lemma Verifier}} have very low FPR but miss many errors.
The baseline \imobench~prompt has the highest FNR among all prompts.

\subsection{FPR and FNR With Prompt Ensemble}
\label{app:additional:fpr-fnr-ensemble}
\begin{figure}[h]
  \centering
  \includegraphics[width=0.8\textwidth]{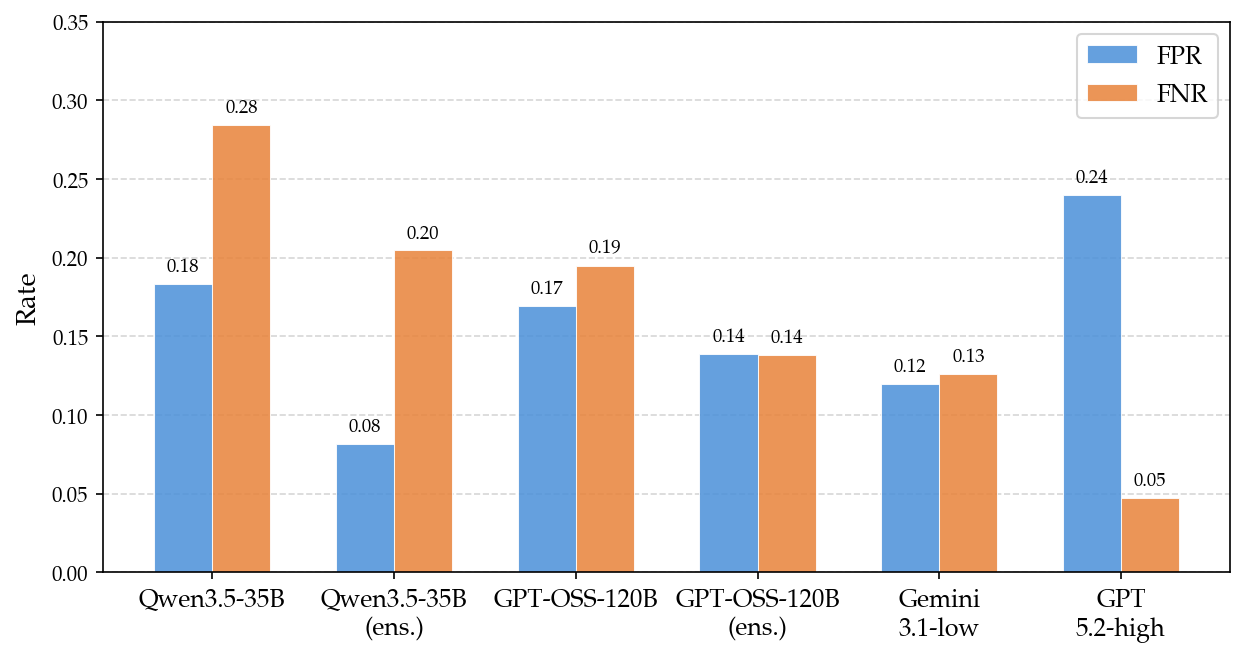}
  \caption{FPR and FNR comparison of open-source models (with and without the prompt ensemble) and frontier models on IMO-GradingBench.}
  \label{fig:fpr-fnr-comparison}
\end{figure}

\Cref{fig:fpr-fnr-comparison} compares the FPR and FNR of open-source models with and without the prompt ensemble against frontier models on the full IMO-GradingBench dataset.
The ensemble substantially reduces FNR for both Qwen3.5-35B and GPT-OSS-120B, bringing them close to frontier-level error detection.
Notably, the ensembled GPT-OSS-120B achieves an FNR of 0.14, matching Gemini-3.1-Low and well below GPT-5.2-High's FNR of 0.05.
The FPR of the ensembled models also remains competitive, with GPT-OSS-120B (ens.) achieving the lowest FPR overall at 0.14.

\end{document}